# Beyond Text: Implementing Multimodal Large Language Model-Powered Multi-Agent Systems Using a No-Code Platform

Cheonsu Jeong[1],*

[1]*Department of AI Automation, SAMSUNG SDS, Seoul, South Korea*

This study proposes the design and implementation of a multimodal LLM-based Multi-Agent System (MAS) using a No-Code platform to address the practical constraints and significant entry barriers in the process of AI technology adoption within enterprises.

**A B S T R A C T**

*Background*: The adoption of advanced AI technologies, such as Multi-Agent Systems powered by Large Language Models (LLMs), poses significant challenges for many organizations due to high technical complexity and implementation costs. No-Code platforms, which enable the development of AI systems without programming knowledge, have emerged as a potential solution to lower these barriers.
*Objective*: This research presents the design and implementation of a multimodal LLM-based Multi-Agent System (MAS) using a No-Code platform, aiming to address the practical challenges of AI adoption in enterprises and demonstrate its effectiveness in automating complex business processes.
*Methods*: The proposed system is designed to process multimodal inputs such as text and images, automate tasks through specialized agents, and reduce the technical burden of AI adoption. Key use cases include code generation from image-based notes, Advanced RAG-based question-answering systems, text-based image generation, and video generation using images and prompts. The system's performance is evaluated through various business application scenarios.
*Results*: The system demonstrated excellent performance in automating business processes. The efficiency of code generation was improved, document search accuracy was enhanced, and image generation time was reduced. These results validate the system's practicality and scalability.
*Conclusion*: The research shows that No-Code platforms can democratize AI technology by enabling general users, not just specialists, to utilize AI for enhancing productivity and efficiency. The multimodal LLM-based MAS framework contributes to lowering AI adoption barriers, promoting its widespread use across various industries.

## 1. INTRODUCTION

The rapid advancement of artificial intelligence (AI) technologies has highlighted the importance of Large Language Models (LLMs) across various fields. Generative AI, which leverages vast amounts of training data to create new content such as text, images, audio, and video, has enabled users to easily utilize generative AI services [1]. Notably, generative AI chatbots have reached a level where they can analyze human emotions and intentions to provide appropriate responses [2], and the advent of LLMs has facilitated their application in tasks such as automated conversation generation and translation [3].

However, LLMs may generate responses that conflict with the most recent information and rely on previously trained data, which limits their understanding of new problems or domains [4]. To address these limitations, various solutions have been explored, including domain-specific fine-tuning of LLMs and utilizing internal knowledge to improve reliability through Retrieval-Augmented Generation (RAG) techniques [5].

Additionally, generative AI technologies based on extensive datasets have made it easier for users to access services that generate new content such as text, images, audio, and video [1], [6]. Multimodal AI models, such as OpenAI's GPT-4V and GPT-4o, demonstrate the capability to integratively process various forms of data, including text, images, and audio [7]. These advancements are driving significant transformations in enterprise environments. Companies are increasingly seeking ways to automate various business processes, such as code generation, RAG-based search, and image processing, using AI technologies. This growing demand underscores the need to enhance productivity and efficiency through AI-driven solutions.

*Address correspondence to the author, Dr. Jeong, at E-mail: paripal@korea.ac.kr or csu.jeong@samsung.com.

Adopting AI technologies in enterprises faces several practical challenges. Building and operating high-performance AI systems require specialized developers and advanced technical resources. Particularly, implementing and managing large-scale AI models such as LLMs involves high complexity, significant costs, and extensive time investments, which makes it difficult for many organizations to easily adopt these technologies [8].

In this context, the concept of No-Code development platforms has recently gained attention. No-Code platforms allow users to build and manage AI systems without programming knowledge, significantly lowering the barriers to AI adoption. According to a report by the Korea Copyright Commission, it is projected that by 2025, approximately 70% of all applications will be developed using Low-Code or No-Code platforms, highlighting the growth potential of this approach. No-Code development environments enhance the accessibility of AI technologies across organizations, enabling not only developers but also non-technical users to leverage AI tools [9].

The objective of this study is to implement a multimodal LLM-based Multi-Agent System (MAS) using No-Code tools and to propose a methodology for effectively integrating AI technologies into business processes without requiring professional developers. To achieve this, the study utilizes workflow-based No-Code platforms such as Flowise to build a system that integrates Multimodal LLMs (MLLMs), image generation, and RAG-based MAS. The feasibility of applying such a system to real-world business processes is evaluated through specific use cases, including analyzing and summarizing captured images for code generation, generating AI-based images, and implementing question-and-answer systems using RAG. Furthermore, the study explores the potential of designing MAS to maximize the specialized roles of agents and create synergies through collaborative operations.

The scope of this study is defined from two perspectives: technical and business processes. From a technical perspective, the study employs Flowise as the No-Code development platform, GPT-4o API for MLLMs, Stable Diffusion for image generation, the Ray model for video generation, and vector databases for RAG implementation. From a business process perspective, the study focuses on applying AI systems to practical environments, including code generation processes, document-based Q&A processes, and image creation processes.

The research methodology consists of three main approaches: literature review, system implementation, and performance evaluation. In the literature review phase, the study analyzes the latest research trends in multimodal AI and MAS and examines case studies using No-Code platforms to establish a foundation for this research. During the system implementation phase, the study designs a No-Code-based system architecture, implements the MAS, and develops the necessary API integrations and user interfaces. Lastly, in the performance evaluation phase, the system's performance is analyzed comprehensively through quantitative evaluations (e.g., processing speed, accuracy, reliability) and qualitative assessments (e.g., user satisfaction, improvement in work efficiency).

This study is expected to provide the following contributions: First, it lowers the barriers to AI adoption, increasing accessibility for non-experts and small-to-medium-sized enterprises. Second, it enhances productivity and operational efficiency by automating business processes using AI and reducing inefficient manual work. Third, it proposes a methodology for building AI systems using No-Code platforms, providing a guideline for AI adoption across various industries. Lastly, it validates the practical applicability of MAS, offering insights into their scalability and future development directions.

## 2. LITERATURE REVIEW

### 2.1 Multimodal Learning
#### 2.1.1 Concept and Advancements of Multimodal Learning

Multimodal Learning is a field of artificial intelligence that emulates the cognitive learning processes of humans by integratively processing and learning from diverse modalities of data [10]. Just as humans acquire and understand information through various senses such as vision, hearing, and touch, multimodal AI systems are capable of simultaneously processing and understanding multiple forms of data, including text, images, audio, and video [11].

The recent integration of multimodal capabilities in major AI models, such as GPT-4o, Claude 3, and Gemini, exemplifies the rapid progress in this field. These models demonstrate not only the ability to recognize images but also the capability to understand and reason about complex relationships between images and text, marking significant advancements in multimodal learning.

#### 2.1.2 Multimodal Fusion Methods

The effective integration of data from different modalities requires robust fusion methods to combine information from diverse sources. Table **1** summarizes the primary multimodal fusion approaches [12].

**Table 1. Multimodal Fusion Methods**

| Fusion Method | Description | Applications |
|---|---|---|
| **Early Fusion** | Combines raw data or low-level features from each modality at the initial stages of processing.<br>· Enables early capture of interactions between modalities, allowing the system to identify cross-modal relationships at the outset.<br>· The unique characteristics of each modality may become diluted during the fusion process. | Video-audio synchronization processing, Emotion recognition systems |
| **Intermediate Fusion** | Processes each modality independently to generate latent representations, which are then fused and processed further.<br>· Enables capturing interactions between modalities at an intermediate stage, allowing for more refined cross-modal relationships.<br>· The additional fusion step increases the system's complexity. | Multimodal sentiment analysis, Video captioning, Image-text alignment |
| **Late Fusion** | Processes each modality independently and combines their outputs (e.g., scores) at the final stage of processing.<br>. Preserves the unique characteristics of each modality effectively.<br>. Struggles to capture complex interactions between modalities. | Multimodal search systems, Cross-modal retrieval systems |
| **Hybrid Fusion** | Combines the advantages of Early, Intermediate, and Late Fusion methods.<br>· Allows for feature extraction and integration at various levels, enhancing flexibility and capturing complex relationships across modalities.<br>· Increases design and implementation complexity. | Autonomous driving (video + sensor data), Multimodal generative models |

### 2.1.3 Multimodal Tasks

Recent multimodal systems are capable of performing various tasks, as shown in Table **2**.

**Table 2. Multimodal Tasks**

| Tasks | Description | Applications |
|---|---|---|
| **Multimodal Question-Answering** | Combines images and text to process complex queries and generate responses based on visual reasoning and understanding.<br>· Enables query comprehension through the combination of image and text.<br>· Visual reasoning and answer generation based on visual elements. | Medical image diagnosis assistance, automated product inspection. |
| **Image-Text Transformation** | Performs tasks that involve the mutual transformation between images and text.<br>· Image Captioning: Describes the content of an image in natural language.<br>· Text-to-Image Generation: Creates images based on textual descriptions.<br>· Visual Question Answering (VQA): Provides answers to questions about an image [13]. | Image Captioning, Text-based Image Generation |
| **Video-Text/Image Transformation** | Performing Bidirectional Conversion Between Text, Images, and Video<br>· Image Captioning: Describes the content of a video in natural language.<br>· Text-to-Video Generation: Creating video content from textual descriptions or prompts | Video Content Search, Smart Shopping, Educational Content Generation |
| **Cross-modal Retrieval** | · Image-to-Video Generation: Creating video content using still images as source material<br>Uses information from one modality to search content in another modality, enhancing search accuracy through semantic matching [14].<br>· Image-based Text Search<br>· Text-based Image Search | Image-based Product Search, Text-based Image Search |

### 2.1.4 Advancements in MLLM Architecture

Recent benchmark results for multimodal LLMs are presented in Table **3**, as introduced in MMMU (Massive Multi-discipline Multimodal Understanding and Reasoning), with the GPT-4o1 model demonstrating capabilities beyond human intelligence [15].

**Table 3. Benchmark for Multimodal LLMs**

| Reset | | | MMMU | |
|---|---|---|---|---|
| Name | Size | Date | Val | Test |
| **Human Expert (High)** | - | **2024-01-31** | **88.6** | - |
| **Human Expert (Medium)** | - | **2024-01-31** | **82.6** | - |
| **GPT4-o1** | - | **2024-09-12** | **78.1** | - |
| **Human Expert (Low)** | - | **2024-01-31** | **76.2** | - |
| InternVL2.5-78B | 78B | 2024-12-05 | 70.1 | 61.8 |
| GPT-4o (0513) | - | 2024-05-13 | 69.1 | - |
| Claude 3.5 Sonnet | - | 2024-06-20 | 68.3 | - |
| Gemini 1.5 Pro (0801) | - | 2024-08-01 | 65.8 | - |
| Qwen2-VL-72B | 72B | 2024-08-29 | 64.5 | - |
| InternVL2.5-38B | 38B | 2024-12-05 | 63.9 | 57.6 |
| Gemini 1.5 Pro (0523) | - | 2024-05-23 | 62.2 | - |
| InternVL2-Pro | - | 2024-07-04 | 62.0 | 55.7 |
| TeleMM | - | 2024-11-18 | 61.4 | 58.2 |
| Llama 3.2 90B | 90B | 2024-09-25 | 60.3 | - |
| GPT-4V(ision) (Playground) | - | 2023-11-27 | 56.8 | 56.1 |

The general architecture of these latest MLLMs is shown in Fig. (**1**) [16].

The general architecture of these latest multimodal language models (MLLMs) can be divided into two main areas: Multimodal Understanding and Multimodal Generation. In the Multimodal Understanding area, the input multimodal data is understood through MLLMs, while in the Multimodal Generation area, desired multimodal data is generated based on prompts.

Moreover, MLLM has three key structural features:

1. **Encoder-Decoder Structure**: Multimodal data is processed through an encoder-decoder framework. In this structure, the Vision Encoder extracts visual features from images, while the Language Encoder extracts linguistic features from text. The Cross-Attention mechanism allows the model to learn the complex relationships between different modalities, facilitating integrated understanding.
2. **Multimodal Embedding Techniques**: MLLMs use multimodal embedding techniques to effectively integrate data from different modalities. This is achieved by creating a unified embedding space that enables semantic alignment across modalities. The integrated embedding space supports efficient cross-modal learning and enhances interactions between various modalities.

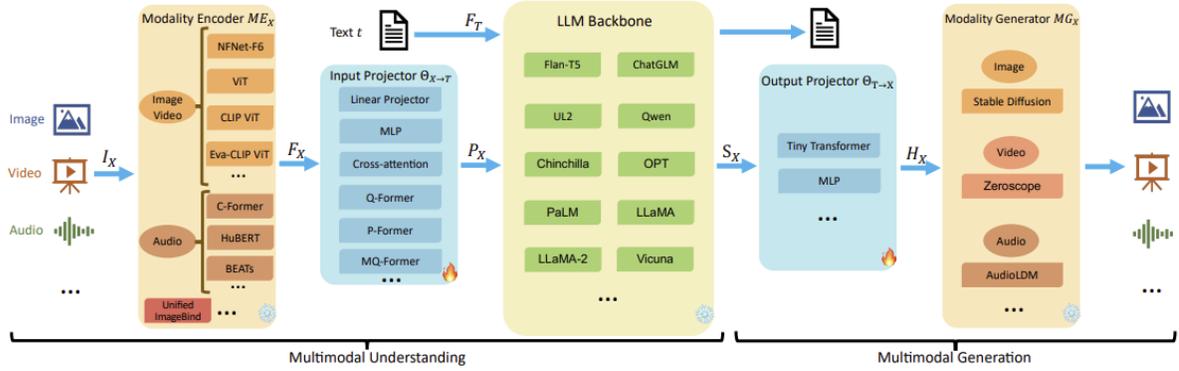

**Fig. (1).** The general model architecture of MLLMs and the implementation choices for each component.

3. **Self-Attention Mechanism**: MLLMs utilize a Self-Attention Mechanism to achieve deep data understanding. This mechanism enables the model to learn relationships not only within each modality but also across different modalities. This strengthens contextual understanding, enabling more refined multimodal processing.

The combination of these structural features allows MLLMs to process and understand diverse modalities of data effectively, enabling the execution of complex multimodal tasks required in modern AI systems.

This content comprehensively covers the basic concepts of multimodal learning and the latest advancements in the field. It emphasizes the practical aspects of each concept and technology, particularly considering their applicability in real-world business environments.

**2.2 Multi-Agent System (MAS)**
**2.2.1 Concept and Structure of MAS**
An agent is a system that uses LLMs to determine the control flow of an application, and it can be categorized into Single Agents, which perform a single task, and Multi-Agents, which work together to perform complex tasks. When the context becomes too complex for a single agent to track, one solution is to divide the application into several small, independent agents and configure. them into a MAS, as shown in Fig. (**2**).

The **Agent Supervisor** plays a central role within the Multi-Agent workflow, coordinating communication and task distribution between different Worker Agents. It receives outputs from an agent, interprets these messages, and then directs the workflow accordingly. The Supervisor is responsible for managing the sequence of tasks, making decisions about the next steps in the workflow, and ensuring the overall efficiency of the system [17].

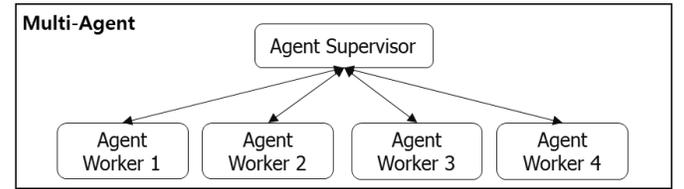

**Fig. (2).** The General Multi-Agent Architecture

A MAS refers to a system in which multiple autonomous agents interact and collaborate to solve complex problems [18]. Each agent possesses independent decision-making capabilities, exchanges information with other agents, and works together to achieve a common goal. With the recent advancements in LLMs, LLM-based MAS are emerging as a new paradigm [19].

Unlike traditional MAS, LLM-based MAS enable more flexible and adaptive collaboration through natural language processing capabilities. Specifically, each agent performs domain-specific tasks while coordinating complex tasks through interactions in natural language [20].

**2.2.2 Characteristics of LLM-based MAS**
The construction of an LLM-based Single Agent system focuses on formalizing interactions between internal mechanisms and the external environment. In contrast, LLM-based MAS emphasize various agent profiles, interactions between agents, and collective decision-making processes, allowing for more dynamic and complex tasks to be solved through the collaboration of multiple autonomous agents, each with unique strategies and behaviors [21].

LLM-based MAS have three key characteristics:

1. **Natural Language-based Interaction**: Communication between agents occurs through natural language interactions. These interactions enable effective information exchange without the need for complex communication protocols, significantly enhancing the system's flexibility [22].
2. **Role Specialization**: Each agent holds a specialized prompt template tailored to a specific task, enabling specialized functionality. This role specialization contributes to improved overall system efficiency [23].

3. **Adaptive Task Allocation**: The system ensures stability and efficiency through an adaptive task allocation mechanism. By considering the workload and capabilities of each agent, dynamic task assignments prevent bottlenecks and enhance overall system stability.

The integration of these features enables LLM-based MAS to efficiently handle complex tasks and provide a flexible, responsive foundation. The harmony between natural language-based interaction, specialized roles, and adaptive task allocation is crucial for improving system performance and scalability.

### 2.2.3 Design Considerations for Agents in Multimodal Processing

The design of a MAS for multimodal processing is based on three core types of agents.

1. **Preprocessing Agents by Modality**: These agents specialize in preprocessing data from each modality, such as images or text, converting it into a standardized format for subsequent processing [24]. This preprocessing ensures that the unique characteristics of each modality are preserved while preparing the data for integrated processing.
2. **Modality Integration Agents**: Based on the Flamingo model proposed by Alayrac et al. [25], the importance of modality integration agents is highlighted. Flamingo, a visual language model (VLM), takes text and visual data as input and generates free-form text as output. These agents play a key role in effectively integrating multimodal information, especially by leveraging cross-modal attention mechanisms to understand and learn complex relationships between modalities. This enables the complementary use of information from different modalities, enhancing understanding and interpretation.
3. **Context Management Agents**: The design of context management agents is crucial for system performance. These agents manage the temporal and spatial relationships between multimodal information, ensuring that consistent responses are generated in a multimodal environment. Effective context management is essential for the overall performance and reliability of the system, particularly when dealing with complex multimodal interactions [26].

For these three types of agents to work synergistically, additional design principles must be considered:
- Efficient communication protocol design between agents
- Data transformation strategies to minimize information loss between modalities
- Parallel processing structures for real-time handling
- Modular architecture design for system scalability

By incorporating these design considerations, a MAS for multimodal processing can provide more effective and stable performance. Specifically, the specialized functions of each agent and their collaborative interaction play a critical role in successfully completing complex multimodal tasks.

## 2.3 Advanced RAG System
### 2.3.1 Concept and Working Principle of RAG

RAG is a hybrid architecture that combines the generative capabilities of LLMs with external knowledge retrieval, addressing the hallucination issue of LLMs and enabling the use of up-to-date and domain-specific knowledge [27]. A RAG system consists of two main components: the Retriever and the Generator. The retriever performs similarity searches based on a vector database, utilizing semantic indexing and search optimization techniques. The generator uses the retrieved context to generate responses, with advanced prompt engineering applied during the generation process.

The operation of the RAG system can be divided into three main stages:

1. **Document Processing Stage**: This includes a series of steps from chunking the original document to storing it in a vector database. Specific tasks include:
   - Deciding optimal chunk sizes for efficient document splitting
   - Using overlap techniques to preserve contextual information
   - Creating embeddings based on the document's characteristics
   - Structuring the vector database for efficient search
2. **Search Stage**: This stage involves converting the user query into a vector space and extracting relevant documents [28]. Multiple search strategies are applied to improve search accuracy, and reranking processes optimize the search results. Dynamic search range adjustments are made based on context.
3. **Generation Stage**: In this stage, the retrieved results and the original query are combined to generate the final response [29]. Contextual appropriateness is ensured using dynamic prompt templates, and the quality of the response is verified for accuracy, consistency, and source traceability via metadata management.

### 2.3.2 Advanced RAG

The Advanced RAG model improves the accuracy and consistency of generated answers by incorporating additional verification and refinement processes. In particular, its multimodal capabilities, which integrate various forms of data, are particularly useful in tasks such as internal document search in organizations.

1. **Multi-Modal RAG**
   Multi-Modal RAG is an innovative information retrieval system that integrates and processes various forms of data, such as text, images, and charts. Unlike traditional text-based search systems, it can generate accurate and relevant responses to complex queries that include visual information. The multi-modal RAG system generates synergy through the interaction of different modalities, simultaneously analyzing both text and visual data to more accurately understand the query's meaning and provide comprehensive related information.

Key advantages of the Multi-Modal RAG system include:

- **Improved Accuracy**: By utilizing information from multiple modalities, the system can more precisely understand the query's intent and provide highly relevant information.
- **Expanded Versatility**: It is capable of processing a wider range of data types, including not just text but also images, videos, and more.
- **Enhanced User Experience**: The system offers intuitive responses to complex queries, improving user convenience.

By integrating multiple modalities, the multi-modal RAG system expands the scope of information retrieval and enhances the information search process.

2. **Agent RAG Application**

   To enhance the RAG process, the Agent RAG model introduces the concept of agents that assist in the answer generation process. By incorporating agents, the model improves the accuracy and consistency of responses [30]. As shown in Fig. (**3**)., agents evaluate the content of the response and perform tasks such as additional web searches to improve the accuracy of the answers.

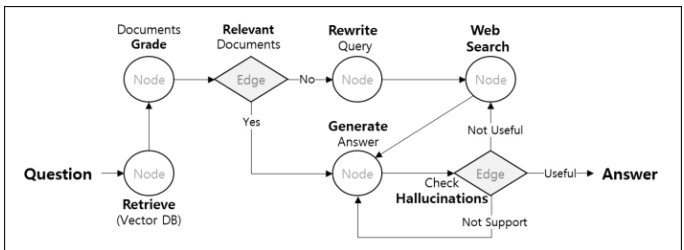

**Fig. (3).** Agent Based Advanced RAG Workflow

3. **RAG in Corporate Document Processing**

   The use of RAG systems in corporate environments has unique requirements and characteristics distinct from general RAG implementations. Specifically, in corporate document processing, the system must effectively handle the diversity and complexity of documents and incorporate the organization's specific domain knowledge.

   Structured document processing is crucial for enhancing data accuracy and usability within corporate documents. Unstructured data, such as tables and charts, is converted into analyzable structures through automated extraction and transformation processes, and metadata within documents is used to maintain context and relevance. Furthermore, various document formats (PDF, DOCX, XLSX) need to be standardized into consistent data formats. Security management is also a key consideration, as corporate documents often contain sensitive information. A detailed access control system must be implemented to manage user access to documents and block unauthorized data access. Additionally, an automatic filtering mechanism for sensitive data (e.g., personal information, confidential data) and an audit trail system to track document access and usage are essential.

   Moreover, RAG systems tailored to the company's domain are required, with a focus on specialized terminology and the integration of organizational context to achieve better outcomes. Managing company-specific jargon and acronyms systematically is essential. A domain-specific dictionary should be constructed and integrated into the RAG system to ensure that the system understands and consistently interprets specialized terms [5]. Additionally, the expansion and management of abbreviations and synonyms are necessary for accurately understanding documents with industry-specific terminology.

   These optimizations in preprocessing and domain-specific handling allow RAG systems to play a critical role in enhancing knowledge management and maximizing operational efficiency in a corporate environment.

**2.4 No-Code Development Platforms**

No-Code development platforms are environments that enable the creation of applications without the need for traditional programming knowledge. These platforms allow non-developers to build simple business applications, while professional developers can focus on more critical tasks such as system architecture and cloud server management by distributing the development workload [3]. The importance of No-Code platforms is increasingly recognized in the AI field, as they significantly contribute to the Democratization of AI technology. These platforms are designed to be easily used by non-experts, enabling rapid prototyping and iterative modifications.

**2.4.1 Advantages and Disadvantages of No-Code Platforms**

1. **Advantages of No-Code Platforms**

No-Code platforms provide notable advantages in terms of development productivity, accessibility, and operational efficiency.

- **Development Productivity**: The drag-and-drop visual development environment eliminates the need for complex coding processes, significantly reducing development time. Additionally, it allows for rapid prototyping using pre-built components, enabling quick responses to market changes.
- **Accessibility**: These platforms allow application development without the need for specialized technical knowledge, enabling non-developers to build complex systems using AI technology, contributing greatly to the democratization of technology.
- **Operational Efficiency**: Three key benefits are observed in terms of operational efficiency:
    - First, needed features can be implemented without the involvement of specialized developers, optimizing IT resources.

- o Second, cloud-based operation improves operational efficiency and reduces costs.
- o Third, integration with a variety of external services through APIs allows for the expansion of application functionality.

Additionally, No-Code platforms offer numerous advantages in utilizing AI technology, including:

- **LLM Integration**: Platforms such as Langflow and Flowise easily integrate with LLMs like GPT-4, Gemini, and Claude.
- **Computer Vision Modules**: Image processing, object detection, and OCR capabilities can be implemented with No-Code tools.
- **Natural Language Processing**: Built-in nodes can be used for tasks like document summarization, sentiment analysis, and question answering.
- **Multimodal Integration**: These platforms can combine and process multiple types of data, including text, images, and audio.

In this way, No-Code platforms play a key role in the democratization of AI and in promoting digital transformation within enterprises.

2. **Disadvantages of No-Code Platforms**

Despite their advantages, No-Code platforms come with several significant limitations.

- **Challenges in Implementing Complex Logic**: A major limitation is the difficulty in implementing complex business logic or mathematical modeling. For projects that require advanced functions or intricate workflows, No-Code platforms may not be sufficient.
- **Limitations in Handling Complex Projects**: These platforms may struggle with handling complex projects that involve extensive workflows or large-scale data manipulation. As the complexity of the application or user base increases, performance bottlenecks may arise, or scalability may become a limitation [31].
- **Customization Restrictions**: One of the primary issues is the limited ability for customization. Since development is based on predefined components and features, implementing specific or complex functionalities can be restrictive.

Considering both the advantages and disadvantages, No-Code platforms are ideal for projects that require quick development and easy accessibility. However, for projects that demand extensive customization or high performance, No-Code platforms should be used with caution.

### 2.4.2 Comparison of Workflow-based AI Multi-Agent No-Code Platforms

Effective implementation of MAS through No-Code platforms requires an intuitive drag-and-drop interface that provides an environment for organically linking various agents around workflows. Currently, open-source workflow-based No-Code AI platforms such as Langflow, Flowise, and n8n are gaining attention, each offering distinct approaches for designing and executing AI workflows. This study compares and analyzes the features of these platforms and discusses their respective strengths and weaknesses to assess their applicability in real-world scenarios.

The selection of a No-Code platform is a key factor in determining the scalability, autonomy, and level of developer control within a system. This section compares three major No-Code platforms with unique characteristics and provides objective criteria for selecting an implementation platform.

1. **Flowise**

   Flowise combines AI models with a visual process designer, enabling users to easily design workflows. Flowise supports various AI models, including LLM, Computer Vision (CV), and NLP, and allows for rapid workflow design through its visual process designer. The platform offers a lightweight runtime environment based on Node.js and Docker support, ensuring stable deployment and integration with various external APIs. A notable strength of Flowise is its multimodal processing capabilities and robust architecture based on TypeScript. Additionally, Flowise enhances scalability and maintainability through its built-in API management, real-time execution monitoring, and plugin extension system [32].

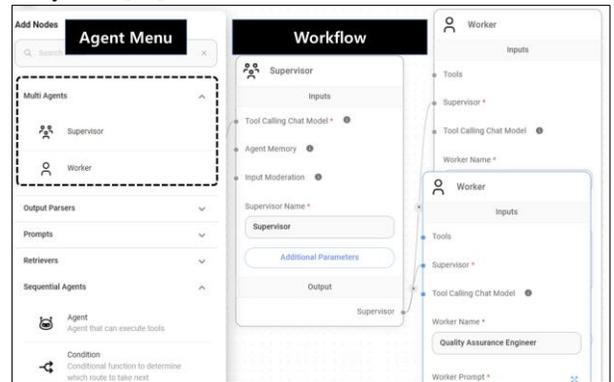

**Fig. (4).** Flowise Agent Workflow

Fig. (**4**). shows the agent workflow in Flowise, where the user's input is processed through a node-based interface. Each node is designed and executed independently, allowing for flexible processing of complex multimodal data.

2. **Langflow**

   Langflow specializes in designing and executing conversational AI workflows and is highly compatible with LangChain components. It offers a scenario editor for creating customized chatbots and supports various AI functions, including LLM, document retrieval, and image generation. Langflow can be deployed in both cloud and on-premises environments and boasts high scalability and modular integration based on Python. The platform benefits from continuous feature improvements supported by the open-source community. However, Langflow has some limitations, including a complex initial installation process, limited multimodal support, and concerns about the stability of its execution environment [33].

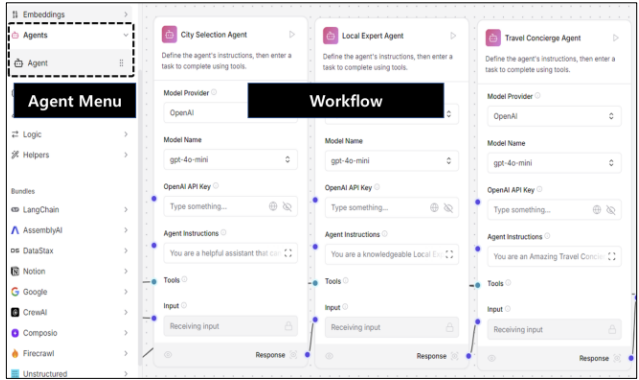

**Fig. (5).** Langflow Agent Workflow

As shown in Fig. (**5**)., the Langflow agent workflow involves sequential execution of various LangChain modules based on user input, allowing dynamic handling of complex workflows.

3. **n8n**

   n8n is a platform offering extensive service integration and powerful workflow automation capabilities. It excels in integrating AI and non-AI tasks, allowing for comprehensive management of workflows involving multiple services. n8n supports enterprise-level features and provides high flexibility for users to control workflows in detail. However, n8n has some limitations, such as a lack of AI-specific features and a steep learning curve, which may pose an entry barrier for novice users [34].

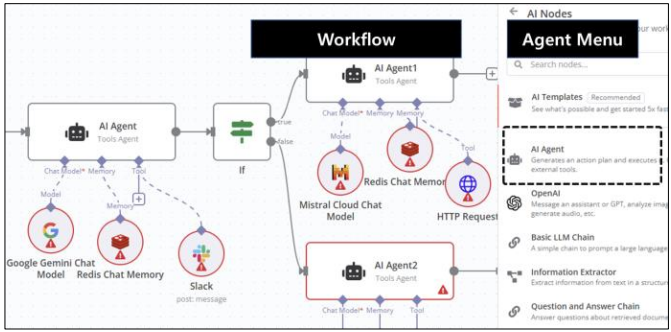

**Fig. (6).** n8n Agent Workflow

The agent workflow in Fig. (**6**). illustrates the process of data integration and workflow execution across multiple services in n8n. The simplified structure is well-suited for automating a variety of tasks, but it is limited in terms of AI-specific features.

**Table 4. Comparison of Workflow-based AI Multi-Agent No-Code Platforms**

| Platform | Key Features | Remarks |
| --- | --- | --- |
| Flowise | · Multimodal processing, intuitive node-based interface<br>· Relatively high initial setup required | Multimodal AI and Multi-Agent Workflow Design |
| Langflow | · Custom chatbot creation, compatibility with LangChain<br>· Limited multimodal support, complex installation | Development of Conversational AI and Chatbots |
| n8n | · Service integration and workflow automation<br>· Lack of AI-specific features, steep learning curve | Data Integration and Non-AI Task Automation |

To build a chatbot that serves as a channel between users and services, it is crucial to select an appropriate chatbot platform in advance and provide functionalities that can immediately respond to user demands [35]. Notably, multi-agent systems can operate by transmitting execution prompts and receiving results through channels such as chatbots. Through the comparison of three platforms presented in Table 4, organizations can select suitable platforms based on their user requirements and operational environment. Flowise demonstrates particular strength in processing multimodal data, making it valuable for applications utilizing diverse data formats. Langflow provides specialized features for conversational AI and chatbot development, making it suitable for environments where user interaction is paramount. n8n serves as a platform capable of integrating both AI-based and non-AI tasks, proving particularly effective in data-centric organizations. The agent flow of each platform intuitively presents various use cases by functionality, which can serve as crucial reference material in platform selection.

## 3. SYSTEM DESIGN AND ARCHITECTURE

### 3.1 Overall System Architecture

The multimodal LLM-based MAS proposed in this study is illustrated in Fig. (**7**)., which adopts a modular architecture designed with scalability and maintainability in mind. The overall system structure is divided into four main layers: Multimodal Input Layer, Multi-Agent Layer, Process Layer, and Multimodal Output Layer.

The Multimodal Input Layer receives various modality inputs such as text, image, and audio. The Multi-Agent Layer consists of the Agent Supervisor and Agent Workers, with the Agent Supervisor managing the collaboration and task distribution among the worker agents. The Agent Workers include RAG, Web Search, Image Analysis, Image Generation, and Audio Generation Agents. The Process Layer houses the MLLM, where each worker agent's characteristics are mapped to the corresponding LLM and functions for processing. Finally, in the Multimodal Output Layer, answers in the form of text, image, video, and audio are provided by each agent based on its specific task.

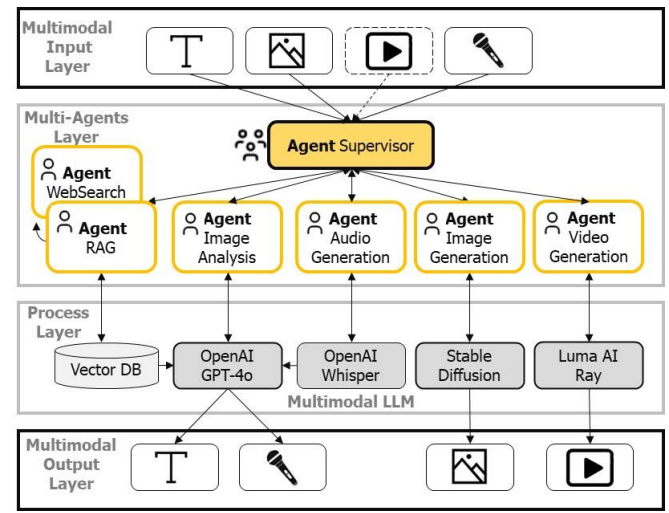

**Fig. (7).** Multimodal LLM-based MAS Architecture

### 3.2 Multi-Agent Definition and Architecture Design

This section proposes the detailed design of the multimodal LLM-based MAS, focusing on core system components such as agent interactions, task distribution, data flow, and external system integration.

### 3.2.1 Agent Types and Role Definitions

The agent structure in this study is divided into two primary categories: Supervisor Agent and Worker Agents. The Supervisor Agent plays the role of the coordinator for the entire system, while the Worker Agents are designed to specialize in specific domains. This structure ensures the autonomy of each agent while maintaining seamless cooperation within the system.

- **Supervisor Agent**: Acts as the central controller of the system, managing task distribution, setting priorities, handling error management, and overseeing the overall system status. One Supervisor Agent oversees multiple Worker Agents.
- **Worker Agent**: These agents perform specialized tasks as directed by the Supervisor Agent. For example, an image analysis agent handles visual information, while a RAG query agent handles question answering.

### 3.2.2 Integration Structure of Multi-Agent System

The integration structure of each agent in the multimodal LLM-based MAS follows a hierarchical structure. The Supervisor Agent and Worker Agents interact within this hierarchy, which ensures clear role allocation and reduces overall system complexity. Each agent operates autonomously within its designated layer, following the directives of the Supervisor Agent to meet the system's collective goals.

### 3.2.3 Collaboration and Task Distribution Among Agents

To ensure efficient communication and task distribution among agents, the system is designed with a communication framework that includes message structuring, priority level definitions, and routing mechanisms. Task orchestration is based on the prioritization of tasks, resource requirement analysis, and dynamic capacity allocation.

1. **Collaboration Mechanism**

   Effective collaboration among different Worker Agents is essential to handle complex tasks. This system adopts the following collaboration strategies:

   - **Information Sharing**: Each agent shares the results or intermediate outcomes of its tasks with others, facilitating the use of this information in other agent tasks. This maximizes synergy and minimizes redundancy.
   - **Role Division**: Each agent focuses on the task it performs best, increasing overall system efficiency. For example, the image analysis agent specializes in visual data processing, while the RAG search agent focuses on text-based information retrieval.
   - **Negotiation**: Agents negotiate priorities, resource allocations, and other factors to find the best possible solution. In this process, each agent's objectives and constraints must be considered to reach a consensus.

2. **Task Distribution Strategy**

   To ensure efficient task distribution, the following strategies are implemented:

   - **Task Decomposition**: Complex tasks are broken down into smaller sub-tasks, which are then assigned to the appropriate agents. This allows for parallel processing and improves system processing speed.
   - **Dynamic Allocation**: Task allocation is dynamically adjusted based on the system's state or task characteristics. For example, if an agent becomes overloaded, tasks may be reassigned to other agents.
   - **Priority-Based Allocation**: Tasks are assigned priorities based on their importance or urgency, with high-priority tasks processed first. This ensures optimal system performance.

3. **Architecture for Collaboration and Task Distribution**

   The architecture for collaboration and task distribution in this system includes the following components:

   - **Centralized Management System**: This system monitors the overall status of tasks, assigns tasks to agents, and coordinates collaboration.
   - **Agent Communication Protocol**: A standardized communication protocol ensures efficient message passing between agents.
   - **Task Queue**: Tasks to be assigned to agents are stored in a queue, allowing for priority management and ensuring that agents receive tasks dynamically when needed.

This architecture ensures effective collaboration and task distribution, enabling the MAS to function efficiently while maintaining flexibility and adaptability in response to changing workloads and priorities.

**3.2.4 Prompt Chain and Agent State Management**

1. **Prompt Chain Design**:

   The Prompt Chain is a crucial structure for AI agents to interact and perform tasks efficiently. The design of the prompt chain is organized in a way that allows tasks to be processed sequentially, incorporating feedback loops to continually improve the results. The chain provides flexibility, enabling different paths based on task conditions to respond to various scenarios. Additionally, the prompt chain maintains task context to ensure consistent results and includes memory buffers and history tracking for temporarily storing and referencing necessary data. Lastly, dynamic parameter adjustments and automatic improvements ensure that the prompt chain delivers optimal task performance.

2. **Agent State Management**:

   Managing the state between agents is an essential design element to maintain system consistency and efficiency. The state of each agent is stored in a central repository to ensure consistent management, with distributed storage reducing the load on the central server. A version control system for state information allows for rollback to previous states, and real-time synchronization and state propagation ensure all agents use the latest data. In case of state conflicts, an automatic conflict resolution mechanism is included, ensuring system stability through a recovery system.

3. **Error Handling and Recovery Strategy**:

   Error handling and recovery strategies are critical for ensuring system reliability. These strategies continuously monitor the system for exceptions, enabling early detection and swift responses to errors. Thresholds are set to immediately alert users upon errors, and an automatic retry mechanism resolves issues. When problems arise along critical paths, alternative paths are activated, and performance is gradually reduced to prevent a complete system halt. Backup system switching and preventive measures minimize data loss and ensure an efficient recovery process, maintaining system stability.

**3.3 Flowise-based MAS Implementation Design**

This study presents a design methodology for constructing a multimodal LLM-based MAS utilizing the Flowise platform, which incorporates functionalities to facilitate the implementation of both Supervisor Agent and Worker Agent concepts. The proposed architecture is designed to enable each agent to operate independently while maintaining effective collaborative capabilities.

**3.3.1 Key Features of Flowise and Multi-Agent Support**

Flowise is based on a modular architecture, ensuring that each component operates independently but is organically connected to implement the system's functionality. The core architecture is built with modular elements, such as the Frontend, API Layer, Core Engine, and LLM Services, ensuring high scalability and maintainability. Additionally, Flowise supports asynchronous node execution, memory-efficient stream processing, and an error recovery mechanism for efficient task processing.

Flowise provides the following key features for implementing a MAS:

- **Agent Templates**: Predefined templates for specific roles, allowing for quick agent creation and customization.
- **State Management**: Synchronizes the state between agents using a distributed state store and manages state changes through transactional mechanisms.
- **Monitoring and Debugging**: Real-time agent monitoring and log tracking for troubleshooting.

Two main structures are used for building a MAS:

- **Chain-based Agent Configuration**: Using the Agentflows category, the Multi-Agents menu in Flowise, the Supervisor and Worker cards are combined to create a hierarchical agent structure. The Supervisor manages and distributes the overall tasks, while the Workers handle specific sub-tasks. Collaboration between agents occurs through message passing between chains.
- **State Management**: Shared memory systems are used to maintain conversation context and synchronize the progress of tasks in real-time.

**3.3.2 Flowise-Based Design and Implementation Procedure**

1. **Create a New Workflow in Agentflows**: Initiate a new workflow for the MAS.
2. **Place the Supervisor Card**: Position the Supervisor card to define the management role of the overall system.
3. **Add Worker Cards**: Add the required number of Worker cards, assigning them specialized roles.
4. **Set Up Message Passing Paths Between Agents**: Establish communication paths between agents for collaboration.
5. **Configure Shared Memory**: Set up shared memory to maintain the conversation context and synchronize task progress between agents.
6. **Define Prompt Templates and Task Rules**: Define each agent's prompt template and task rules to control agent behavior.
7. **Multi-Agent Integration in the Chatflows Category**: To efficiently operate multiple agents within a single channel, an integrated workflow is created. In this context, the 'Chatflow Tool' is employed to enable various agents to interact seamlessly, ensuring that they work harmoniously within a unified flow.

The Flowise platform offers a modular architecture with various features that provide an ideal environment for implementing a MAS. This section detailed how to implement a MAS using Flowise, offering developers a comprehensive method for efficiently designing and building complex MAS.

## 3.4 Use Case-based MAS Implementation Design

### 3.4.1 Image Analysis and Code Generation Agent: Image to Text, Text Code to Code

This agent analyzes the input image and generates the optimal, completed code based on the analysis results, which are prompted by the image.

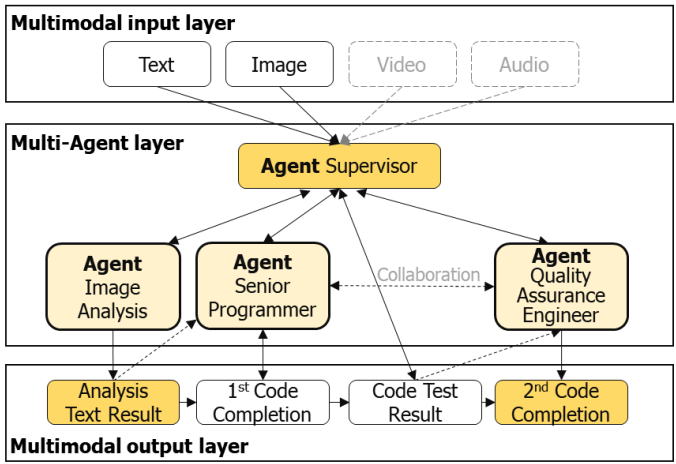

**Fig. (8).** Image Analysis and Code Generation Agent Architecture

The agent analyzes a sketch of the sample code image, generates the incomplete code into a complete version, and involves a Quality Assurance role to finalize and enhance the code. The input is primarily an image upload, and additional text input specifies the desired outcome. The architecture includes the Supervisor Agent, Senior Programmer Agent, and Quality Assurance Engineer Agent, who collaborate to achieve the target (Fig. (**8**)).

1. **Image Capture Processing**

   The first step of code image analysis is capturing the handwritten code as an image and converting it to text. This process includes adjusting the image size and resolution to better recognize the information within the image. During preprocessing, the important parts of the code are highlighted, the text's clarity is adjusted, and unnecessary areas are removed, increasing the accuracy of text recognition and automating the analysis of the code content.

2. **Text Extraction and Structuring**

   Optical Character Recognition (OCR) is used to extract text from the image, converting all textual information within the image into digital data. The extracted text is structured into a consistent format, categorized into items like speaker, topic, and keywords. This structure facilitates quick identification of key terms during the code image analysis phase and improves the efficiency of summarization and searching.

3. **Code Accuracy Evaluation**

   To evaluate the accuracy of the image capture and text extraction processes, comparison data is used to analyze the matching degree between the extracted information and the original code, as well as to verify the generated code by running it. The accuracy of text recognition depends on the OCR model's performance, which may vary depending on image characteristics like background, font size, and resolution. Improvements can include enhancing preprocessing with adjusted image resolution and contrast or applying the latest OCR models or various MLLMs for better performance.

### 3.4.2 RAG Search Agent: RAG to Text, Audio to Text

Based on Advanced RAG [30], the RAG search agent performs three core functions: text-based query processing, document search, and response generation. If the RAG model cannot find an answer, the Web Search Agent generates a more accurate response [30]. Inputs can be in both text and audio, with audio inputs converted to text using Speech-to-Text technology (Fig. (**9**)).

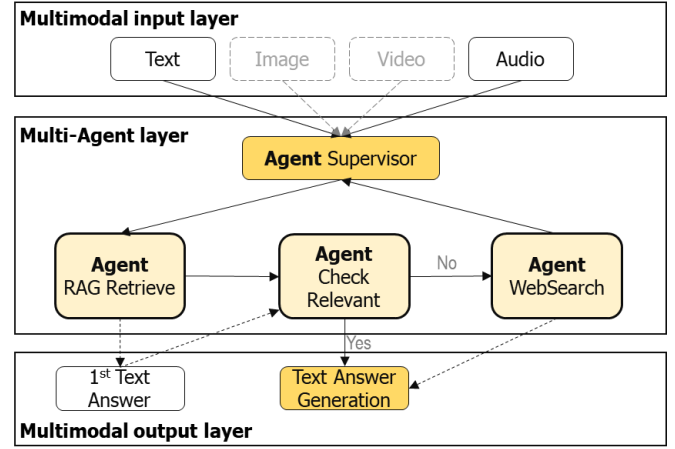

**Fig. (9).** RAG Agent Architecture

The agent plays a crucial role in the final answer generation process:

- **Answer Evaluation**: Evaluates the accuracy, fluency, and reliability of the answers generated by the RAG model in the first stage.

- **Answer Improvement**: Based on the evaluation results, the answer is refined.

- **Information Retrieval**: Additional information is searched to improve the answer.

### 3.4.3 Image Generation Agent: Text to Image, Image to Image

The 'Text to Image' form generates related images based on a prompt describing the image, while the 'Image to Image' form uses an image-specific model to sketch an image similar to the target and then submits it as input for the agent to analyze and generate the optimal image (Fig. (**10**)).

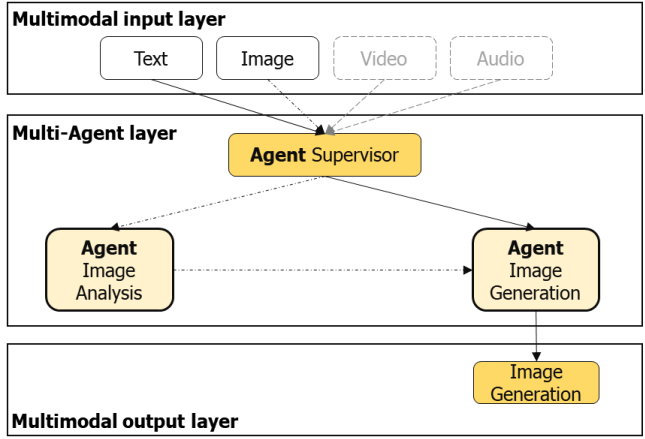

**Fig. (10).** Image Generation Agent Architecture

The image generation agent uses the Stable Diffusion model to generate unique images based on the given prompt.

1. **Prompt Optimization Process**

   The image generation process involves creating an image through the Stable Diffusion model based on the provided prompt. Prompt optimization involves refining the user's desired style, color, and details to clarify the goal of image generation. In this step, various input elements are adjusted to ensure that the model generates specific, high-quality images. The prompt optimization process is repeated to reflect user requirements and achieve the best possible outcome.

2. **Generated Image Quality Evaluation**

   The generated image is evaluated based on criteria like resolution, color accuracy, and detail implementation. Both subjective and objective assessments are used to evaluate the quality, with particular attention to how well the generated image matches the prompt. This helps assess the performance of the image generation model, and if necessary, adjustments are made to the prompt or generation parameters to improve quality.

### 3.4.4 Video Generation Agent: Text to Video, Image to Video

The 'Text to Video' form is designed to generate videos based on prompts describing the desired content, leveraging the capabilities of MLLM. The 'Image to Video' form utilizes an image-specialized model, where the user provides a sketched or original image similar to the target video as input. The agent analyzes the image and references the target prompt to generate the optimal video (Fig. (**11**)).

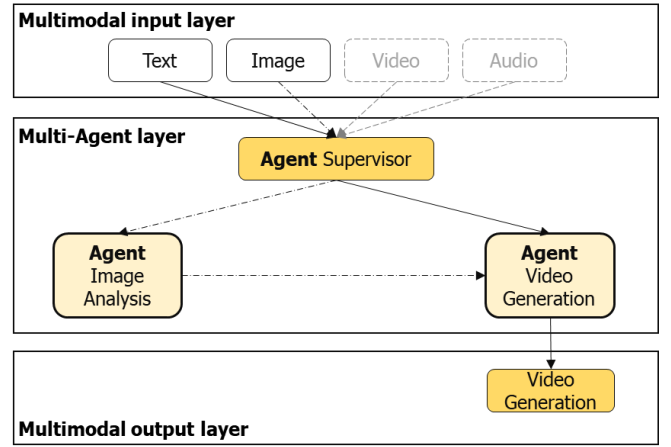

**Fig. (11).** Video Generation Agent Architecture

The video generation agent employs the Ray model by Luma AI to create unique videos based on the given prompts. The video generation process involves a prompt optimization phase, where the Ray model generates videos according to the provided prompts. Prompt optimization focuses on clarifying the goals of video generation by specifying aspects such as style, color, and detail. During this phase, various input elements are adjusted to enable the model to produce detailed and high-quality videos. The prompt optimization process is iteratively refined to achieve results that best meet user requirements.

## 4. SYSTEM IMPLEMENTATION

This chapter presents the setup and detailed implementation of a multimodal LLM-based MAS using the Flowise platform, based on the MAS design introduced in Chapter 3. The implementation is demonstrated through specific case examples.

### 4.1 Development Environment Setup

#### 4.1.1 Flowise Cloud and API Key Setup

To implement the multimodal LLM-based MAS, the first step is to set up the Flowise platform in the cloud environment. Flowise is a No-Code platform that facilitates the easy integration and management of various agents, making it ideal for building complex AI systems that utilize multimodal data and multiple agents. By setting up Flowise in the cloud, users can perform large-scale computations without relying on local resources, and easily integrate and process data across various agents.

The setup process in the cloud involves initializing the server environment and configuring the network to create a stable and scalable infrastructure. Users can allocate resources through their cloud account and deploy the Flowise server to secure the necessary computing power.

For MAS implementation, ensuring secure access to data and external service integration is essential. API authentication and key management are crucial to maintaining the system's security. This ensures that external models, such as OpenAI and Stable Diffusion, are securely connected. API keys serve as an important authentication method to allow communication with external services, and their leakage can pose a risk of data breaches. Therefore, Flowise uses environment variables to configure. API keys, ensuring that only necessary keys are loaded into the system, minimizing exposure to sensitive information. These security settings help ensure the stable and secure operation of the MAS.

Fig. (**12**). illustrates the interface for configuring the keys of various systems used in this study. Specifically, image generation utilizing the image-specialized Stable Diffusion model and video generation employing the Ray model were implemented using the Replicate API platform, which enables the execution and adjustment of open-source AI models and the large-scale deployment of customized models.

### 4.1.2 Database Configuration and External Service Integration

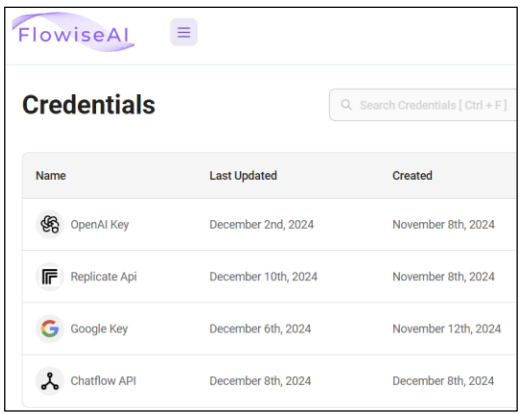

**Fig. (12).** Integration Module Key Configuration

In the Flowise-based MAS, the database plays a crucial role in managing data exchange and state management between the system components. Each agent, such as the image analysis, code generation, RAG search, and image generation agents, generates or consumes unique data, so a robust database is required to effectively store and manage this data. When configuring the database, it is important to record each agent's task history and optimize data access speeds by setting up indexes.

Given the multimodal nature of the data, which includes images, text, documents, etc., the system should be capable of storing and retrieving various data formats. A combination of a relational database for structured data and a NoSQL database for unstructured data is an effective approach. This hybrid database configuration contributes to optimizing the MAS's performance and ensuring data consistency.

To improve the performance and expand functionality, external service integration is necessary. The Flowise platform supports the integration of various external APIs, enabling the use of multimodal AI services and LLMs for knowledge retrieval. For example, integrating OpenAI's API for natural language processing and the Stable Diffusion API for image generation allows the system to process diverse input data in real-time. When integrating external services, it is important to manage network and computational resources based on API call frequencies and processing capacities. Flowise's integrated modules help manage connections with each API, ensuring efficient service and resource allocation.

### 4.2 Multi-Agent Implementation

#### 4.2.1 Image Analysis and Code Generation Agent Implementation

The MAS handles the modality processing of 'Image to Text' and 'Text Code to Code'. For input image analysis, the MLLM, OpenAI's GPT-4o, is applied to analyze the input image and generate text-based responses based on the given prompts for the desired output. Using the Multi-Agent functionality provided by the platform, the system is

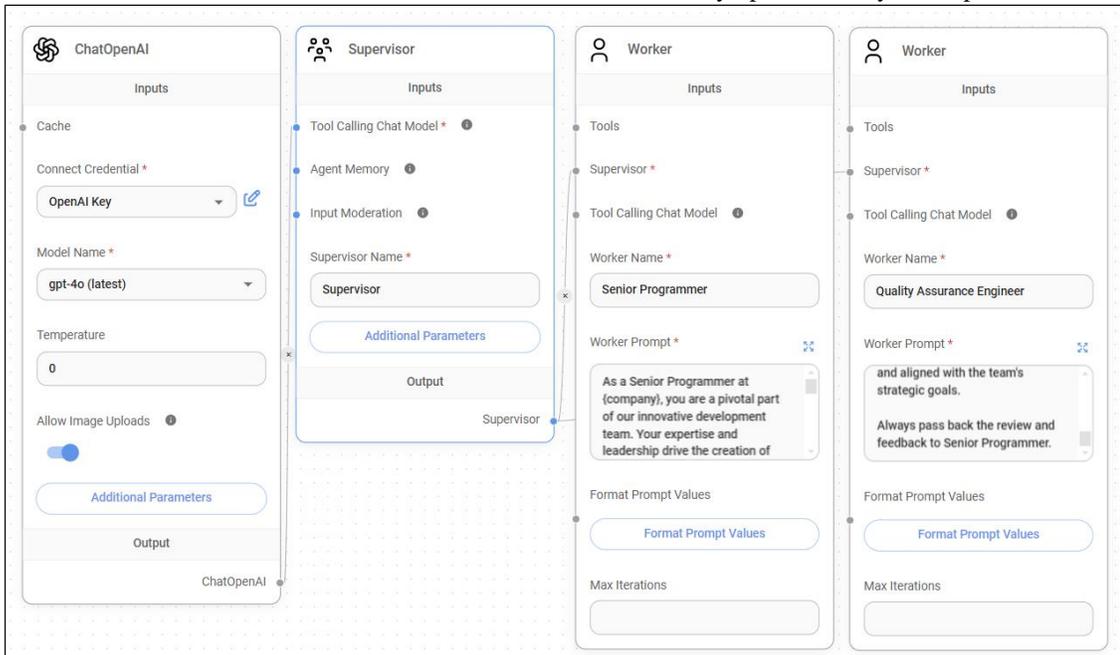

**Fig. (13).** Image Analysis and Code Generation Multi-Agent Workflow Design

structured with Supervisor and Worker Nodes, as shown in Fig. (**13**)., creating the Multi-Agent Workflow. The main prompts used in this implementation are as follows:

***Supervisor Agent System Prompts:***
*You are a supervisor tasked with managing a conversation between the following workers: {team_members}.*
*Given the following user request, respond with the worker to act next. Each worker will perform a task and respond with their results and status.*
*When finished, respond with FINISH. ~ ~ ~ ~*

***Senior Programmer Agent System Prompts:***
*Your goal is to lead the development of high-quality software solutions.*
*The output should be a fully functional, well-documented feature that enhances our product's capabilities. Include detailed comments in the code. Pass the code to Quality Assurance Engineer for review if necessary. Once their review is good enough, produce a finalized version of the code. ~ ~ ~*

***Quality Assurance Engineer System Prompts:***
*Your goal is to ensure the delivery of high-quality software through thorough code review and testing. Review the codebase for the new feature designed and implemented by the Senior Software Engineer.*
*Always pass back the review and feedback to Senior Programmer. ~ ~ ~*

The Worker Agents are composed of the Senior Programmer and Quality Assurance Engineer, and the workflow is implemented as follows. The Supervisor Agent analyzes the sample code image and then directs the Worker Agents to create the completed code. First, the Senior Programmer Agent completes the basic code, and then the Quality Assurance Engineer Agen**t** conducts a code review to provide the final, refined version of the code.

### 4.2.2 RAG Search Agent Implementation

The MAS handles the modality processing of 'RAG to Text' and 'Audio to Code'. Queries are made through Text or Speech for RAG search. To analyze the text, the OpenAI GPT-4o model is used, while GPT-Whisper is applied to recognize speech, enabling the system to process both text and audio inputs. For the output, stored content is searched, and the system generates a text-based response based on the desired prompt. Using the Multi-Agent functionality provided by the platform, the Supervisor and Worker Nodes are employed to structure the Multi-Agent Workflow, as shown in Fig. (**14**)., with the following key prompts being used:

***Supervisor Agent System Prompts:***
*You are a supervisor tasked with managing a conversation between the following workers: {team_members}.*
*If the question is not related to the dress code, have the worker perform a "Web Searcher".*
*When it finds relevant content, it will finish without running any other workers.*

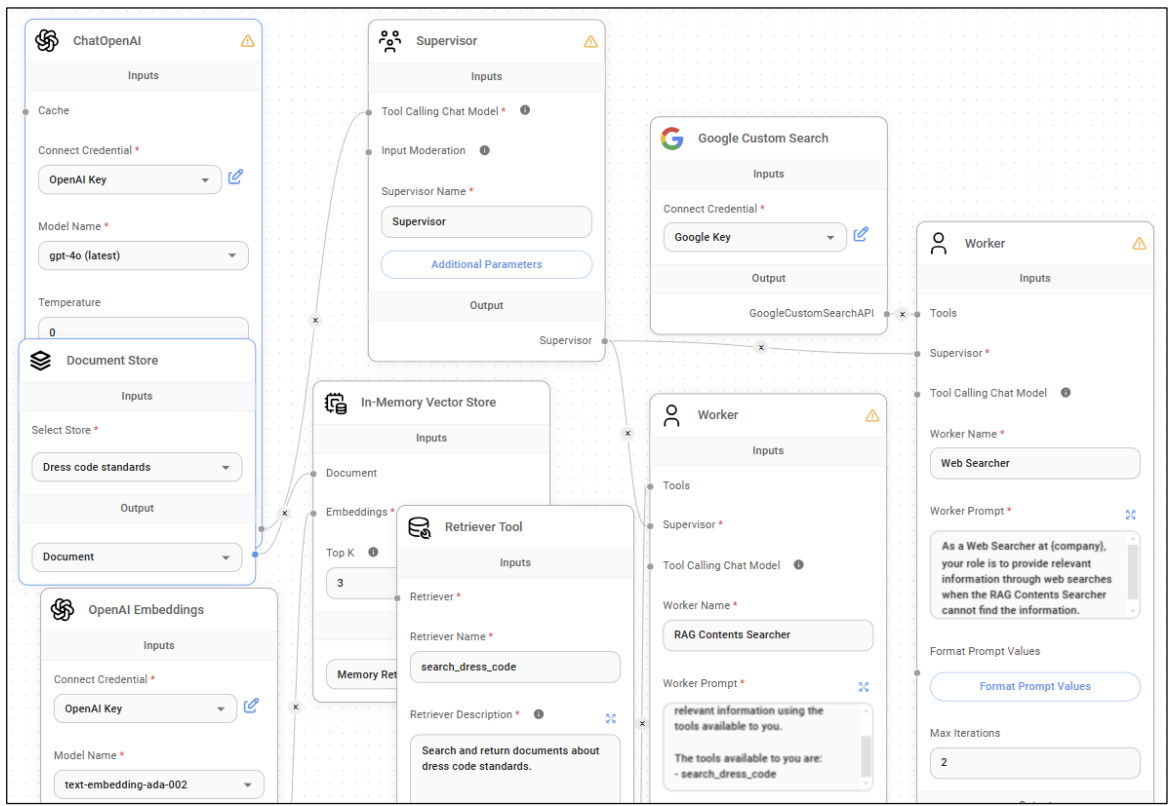

**Fig. (14).** RAG Multi-Agents Workflow Design

*RAG Contents Searcher Agent System Prompts:*
*You are a worker who always answers questions with the most relevant information using the tools available to you.*
*The tools available to you are: search_dress_code*

*Web Searcher System Prompts:*
*As a Web Searcher at {company}, your role is to provide relevant information through web searches when the RAG Contents Searcher cannot find the information.*

The knowledge used for RAG search is processed using the Recursive Character Text Splitter, and the repository is managed through an In-Memory Vector Store. In the RAG Contents Searcher, the system first queries the RAG repository to check if relevant content is available. If content is found, the system generates the final response. If no relevant content is found, the Supervisor Agent directs the Web Searcher Agent to search the web for an appropriate answer, allowing the system to provide responses even for questions not contained in the RAG information. This design ensures that the system can handle queries with both internal and external sources of information.

### 4.2.3 Image Generation Agent Implementation

The MAS handles the modality processing of 'Text to Image' and 'Image to Image'. The system accepts inputs in two forms: Text or Image. For the output, the Stable Diffusion model, which is specialized in image generation, is used. The 'Text to Image' form generates related images based on the provided prompt describing the image, while the 'Image to Image' form first sketches a similar image to enhance the completion quality, and then submits the input to the agent. The agent analyzes the submitted image and, referencing the target prompt, generates the optimal image.

*Query Prompts:*
*{query}*
*Create a high-resolution, clear image with a coherent and logical scene composition. Apply advanced lighting techniques to achieve a dramatic effect. Incorporate fine details and intricate textures, balancing realism with artistic flair suitable for the subject matter.*

*Image Generate Agent Prompts:*
*Generate images tailored to various requirements upon request. Accurately capture the customer's needs and provide customized images that reflect the desired style, theme, colors, and mood.*

When a query about image generation is received, the Query Prompts are referenced to consider factors like image resolution and quality. The system then passes this information to the Image Serve Agent to prepare for image generation. Based on the Image Generate Agent Prompts, the system generates the final image. The major prompts used for this process are as follows:

First, in the 'Text to Image' form, a predefined prompt is referenced to generate the desired image based on a description provided by the user. Second, in the 'Image to Image' form, an input image is given, noise is added to the image, and then the noise is removed to produce the final result. The image is referenced from a sketch similar to the desired output, and based on the prompt, the system generates the optimal image. This process is implemented using the Stable Diffusion model, specifically the latest stability-ai/stable-diffusion-3.5-large-turbo model, which generates the image according to the prompt. The agent is structured as shown in Fig. (**15**). to achieve this.

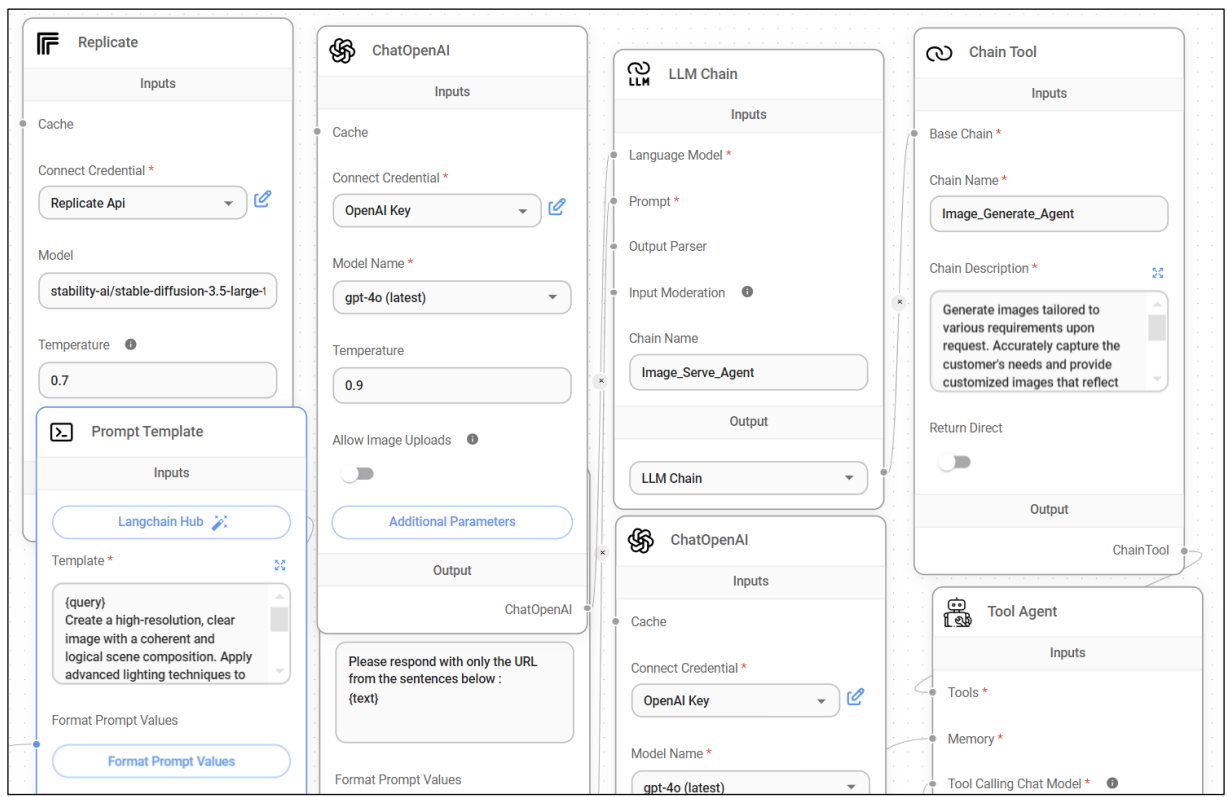

**Fig. (15).** Image Generation Workflow Design

### 4.2.4 Video Generation Agent Implementation

The Multi-Agent system handles the modality processing of 'Text to Video' and 'Image to Video'. Inputs are accepted in two forms: Text or Image, and the outputs are generated using the Ray model, which specializes in video creation.

In the 'Text to Video' form, a prompt describing the desired video is provided, and the system generates a relevant video accordingly. In the 'Image to Video' form, a sketched or original image resembling the target video is submitted as input. The agent analyzes the submitted image and, referencing the target prompt, generates the optimal video.

When a query related to video generation is received, the system references the Query Prompts to gather information about the desired video. This information is passed to the 'Video_Serve_Agent', which prepares the video for generation. Finally, the video is created based on the 'Video_Generate_Agent' Prompts. The main prompts used in this process are as follows:

*Query Prompts:*

*{query}*

*Create a high-resolution, clear video with a coherent and logical scene composition. Apply advanced lighting techniques to achieve a dramatic effect. Incorporate fine details and intricate textures, balancing realism with artistic flair suitable for the subject matter.*

*Video Generate Agent Prompts:*

*Generate videos tailored to various requirements upon request. Accurately capture the customer's needs and provide customized images that reflect the desired style, theme, colors, and mood.*

The 'Text to Video' form generates videos by referencing predefined prompts based on the user's description of the desired video. In contrast, the 'Image to Video' form uses the provided image as input and employs deep learning generation models (e.g., Diffusion models) to naturally create intermediate frames. Additionally, it references sketches or images similar to the desired video and generates the final optimized video according to the prompt. This implementation utilizes Luma AI's latest luma/ray model and is structured as shown in the agent architecture in Fig. (**16**).

## 4.3 Multi-Agent Integration and UI Implementation

### 4.3.1 Multi-Agent Integration Implementation

The four types of MAS—'Image Analysis and Code Generation Agent', 'RAG Search Agent', 'Image Generation Agent', and 'Video Generation Agent'—which were previously implemented, have been integrated into a unified Multi-Agent workflow as shown in Fig. (**17**). Each agent can be executed individually for different user prompts, or all Multi-Agents can be executed together to handle various inputs simultaneously, allowing for a more flexible and efficient response to multiple queries at once.

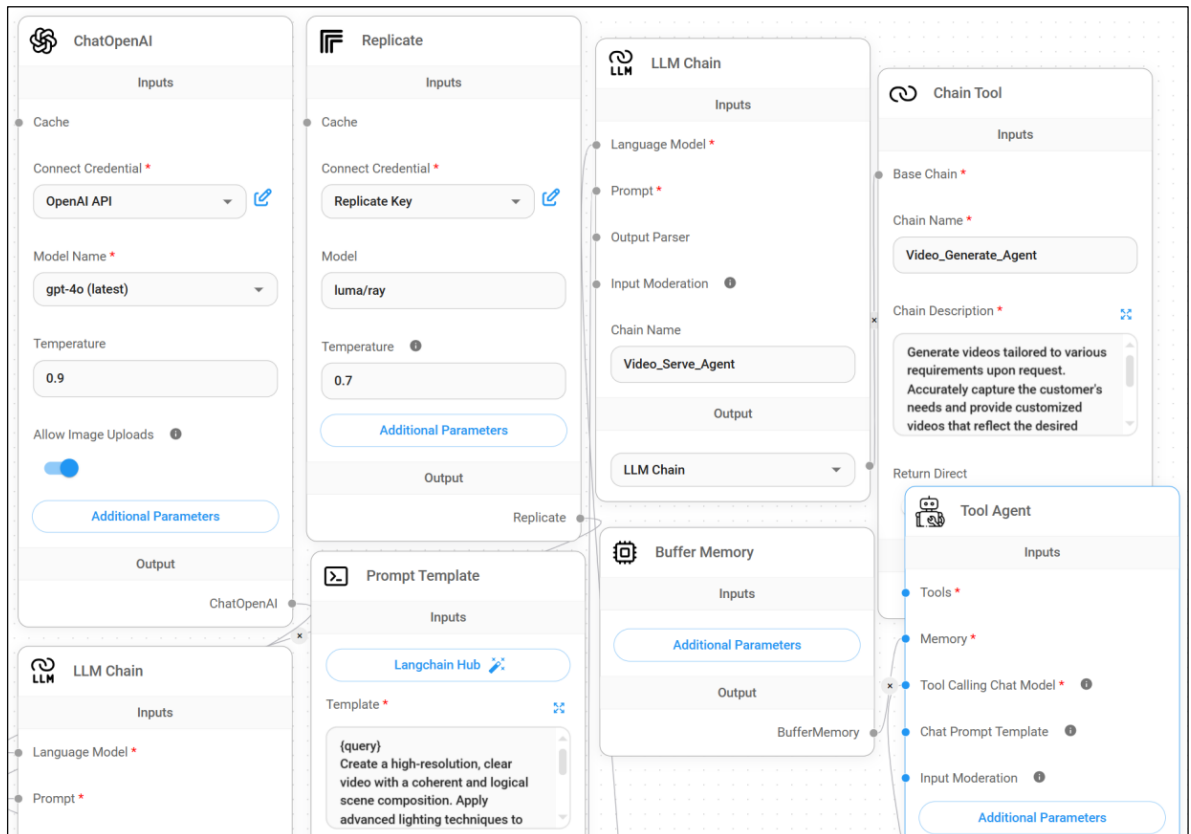

**Fig. (16).** Video Generation Workflow Design

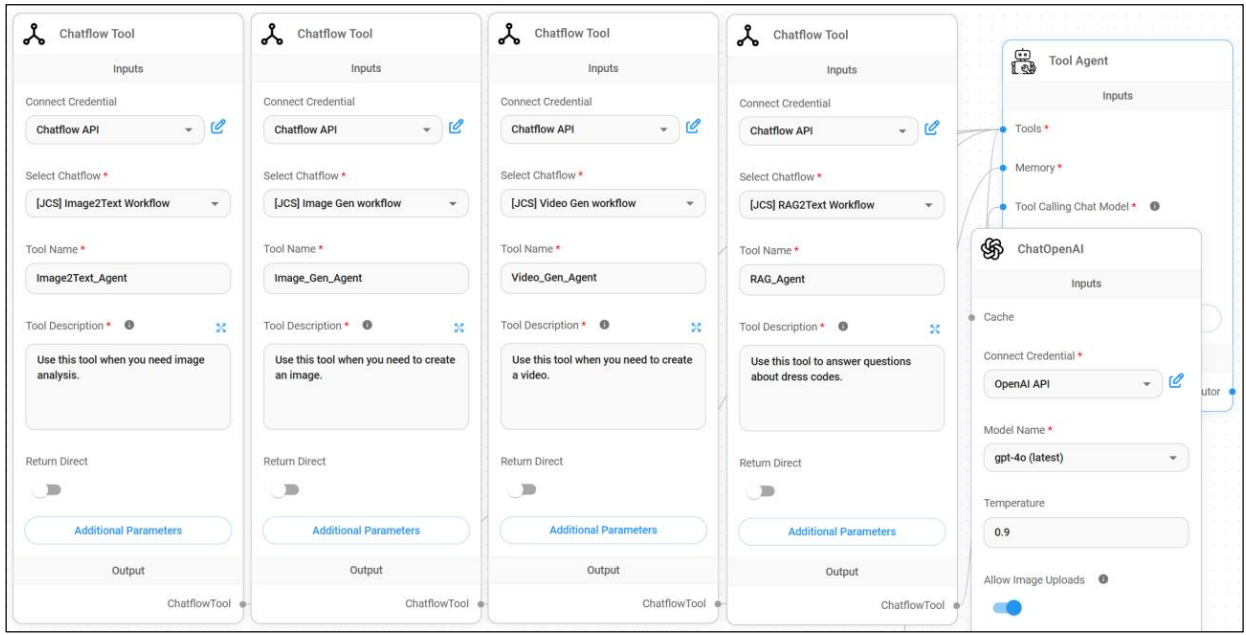

**Fig. (17).** Agent Integration Workflow Design

### 4.3.2 User Interface Implementation

The User Interface (UI) for the MAS is designed to efficiently control the functions of each agent. An intuitive UI allows users to easily perform image analysis, summarization, image generation, and RAG-based searches. The interface offers separate menus and options for each function, ensuring seamless user experience. The UI is web-based, increasing accessibility and enabling real-time user input to be reflected in the system and generate results quickly (Fig. (**18**)).

## 5. IMPLEMENTATION USE CASES AND RESULTS

This chapter analyzes the practical implementation cases of the MAS. Through use cases such as image analysis, code generation, RAG-based search, image generation, and video generation, the performance of the system is evaluated, demonstrating its ability to handle complex tasks through collaboration between agents. The technical performance and practicality of the system are assessed through specific execution results and screenshots.

### 5.1 Image Analysis and Code Generation Use Case

This case demonstrates the implementation of a MAS for analyzing and completing a sketched code image. The system analyzes an incomplete code image, and through collaboration between agents, the code is completed. Afterward, the agents conduct a final quality review to provide the completed code.

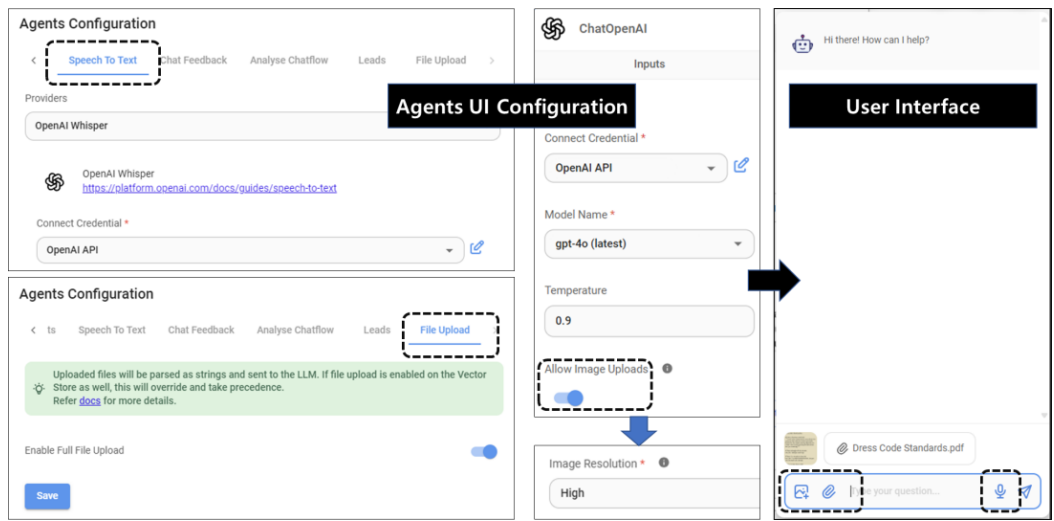

**Fig. (18).** User Interface Implementation Screen

**Fig. (19).** Sketch Code Image (*sample_code.png*)

By registering an incomplete code image like the one in Fig. (**19**). and inputting the prompt "*Analyze the image and complete the code*", the Supervisor Agent analyzes the code and directs the completion of the relevant code, as shown in Fig. (**20**). Upon receiving the instruction, the Senior Programmer Agent completes the code.

**Fig. (20).** Image Analysis Agent Implementation Case

Once the code is completed, the Supervisor Agent directs the Quality Assurance Engineer to review the generated code. After the code review, the Quality Assurance Engineer refines the code based on the results, leading to the final completion of the code, as shown in Fig. (**21**).

**Fig. (21).** Code Generation Agent Implementation Case

## 5.2 RAG Search Query and Answering Use Case

In this case, a question related to RAG knowledge, "*What is the men's dress code?*", was answered by the RAG Contents Searcher Agent, which used the 'search_dress_code' tool to find the relevant content from RAG knowledge. Additionally, a voice query, "*What is the population of South Korea in 2024?*", which was not available in the RAG content, was automatically converted to text (Speech to Text) and forwarded to the system.

**Fig. (22).** RAG Multi-Agent Implementation Case

Through the 'Web Searcher Worker', the 'google-custom-search' tool was used to search Google and provide an accurate answer based on the results. By applying the Advanced RAG model with Multi-Agent functionality, the system was able to enhance the quality of responses even when the original RAG knowledge did not have the answer (Fig. (**22**)).

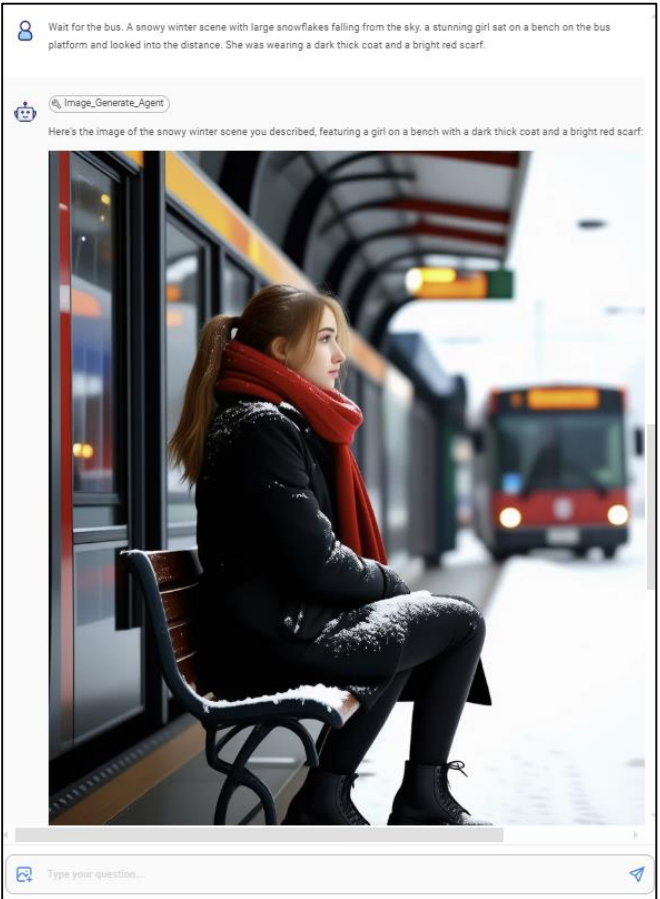

**Fig. (23).** Text to Image Implementation Screen

### 5.3 Image Generation Use Case

The first example is the Text to Image implementation case. When a description of the image to be generated is entered as shown in the user prompt below, the 'Image_Generate_Agent' generates the relevant image, as shown in Fig. (**23**).

*Image Generation User Prompts:*

*Wait for the bus. A snowy winter scene with large snowflakes falling from the sky. a stunning girl sat on a bench on the bus platform and looked into the distance. She was wearing a dark thick coat and a bright red scarf.*

The second example is the Image to Image implementation case. By registering a sketched image file, 'sample mountain.png' (Fig. (**24**))., and entering the user prompt "*Create an image of a fantastic landscape*", the 'Image_Generate_Agent' generates an image similar to the registered one, as shown in Fig. (**25**).

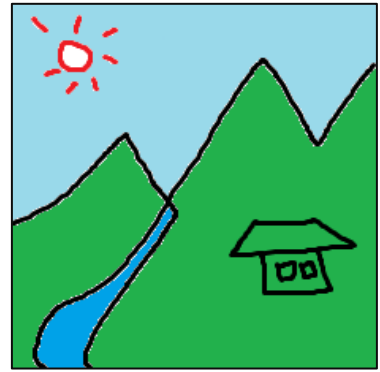

**Fig. (24).** Sketch Image (*sample mountain.png*)

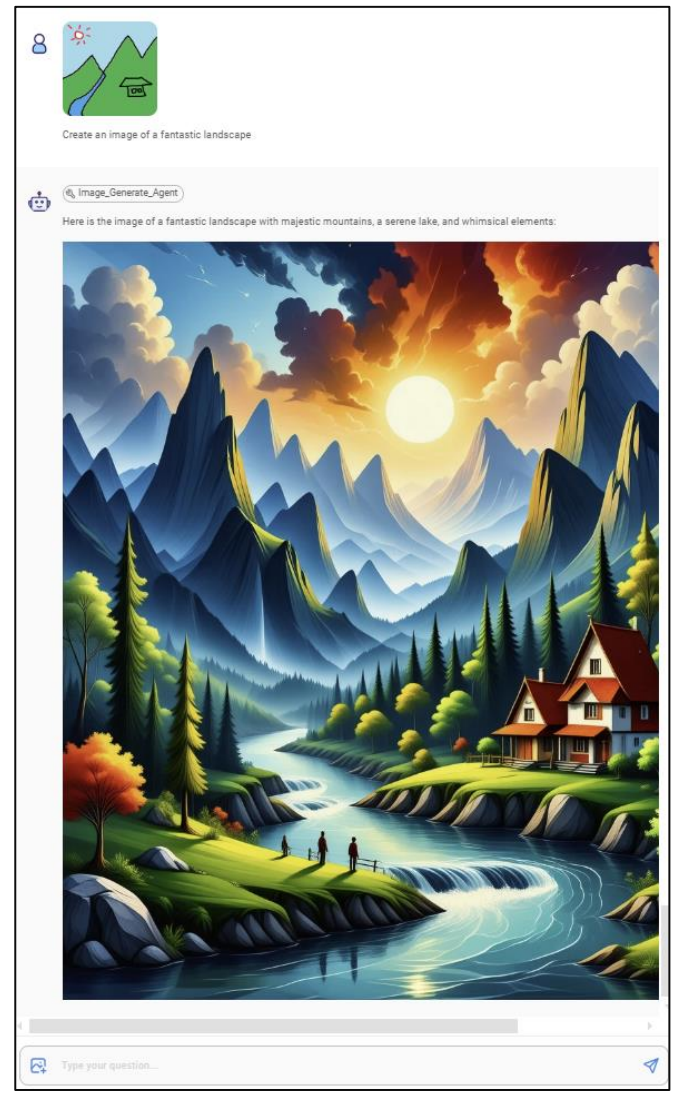

**Fig. (25).** Image to Image Implementation Screen

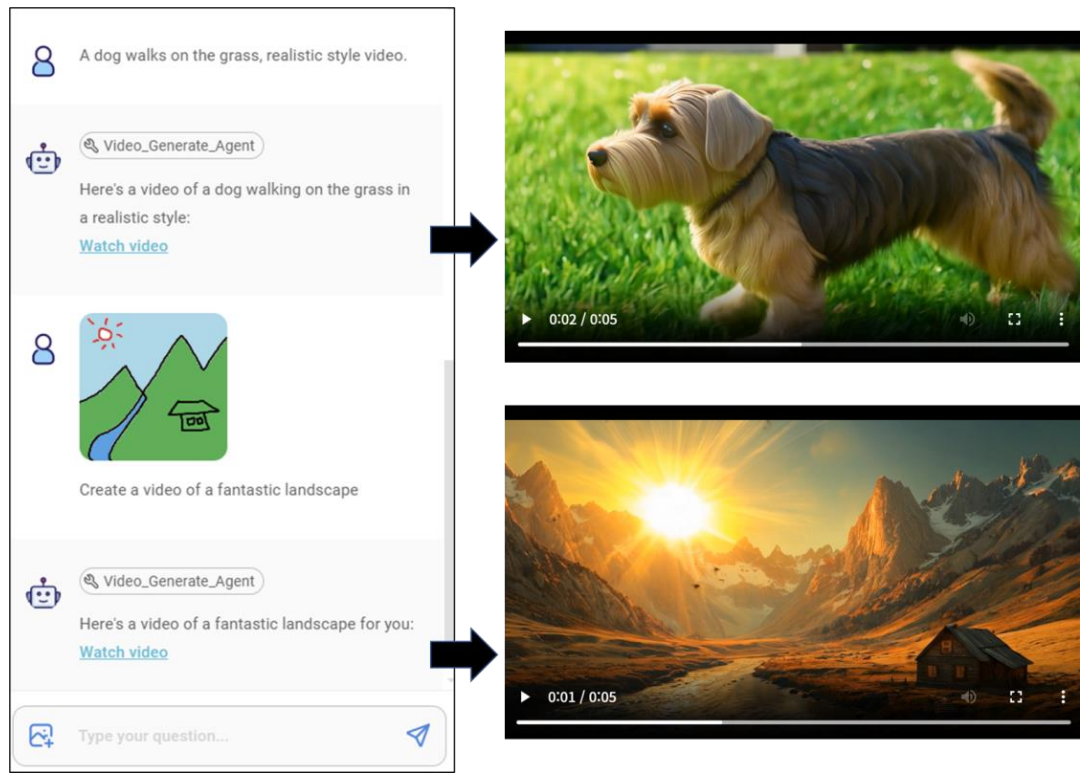

**Fig. (26).** Text/Image to Video Implementation Screen

### 5.4 Video Generation Use Case

The first use case demonstrates the implementation of a Text-to-Video generation system. When the user inputs a prompt, such as "*A dog walks on the grass, realistic style video*", the 'Video_Generate_Agent' successfully generates a video corresponding to the description.

The second use case involves the implementation of an Image-to-Video generation system. A sample sketch image file, such as 'sample mountain.png' (Fig. (**24**))., is uploaded, along with a user prompt like "*Create a video of a fantastic landscape.*" The 'Video_Generate_Agent' then generates a video similar to the uploaded image, as shown in Fig. (**26**).

### 5.5 Multi-Agent Integration Use Case

This case demonstrates the integration of four previously implemented Multi-Agent types—Image Analysis and Code Generation Agent, RAG Search Agent, Image Generation Agent and Video Generation Agent—into a unified MAS. The integrated system allows users to interact with all functionalities through a single UI. For instance, when a question about dress code is asked, the system uses the RAG Agent to provide an answer. Similarly, when a prompt for image generation is entered, the system utilizes the 'Image_Gen_Agent' to generate the desired image, as shown in Fig. (**27**).

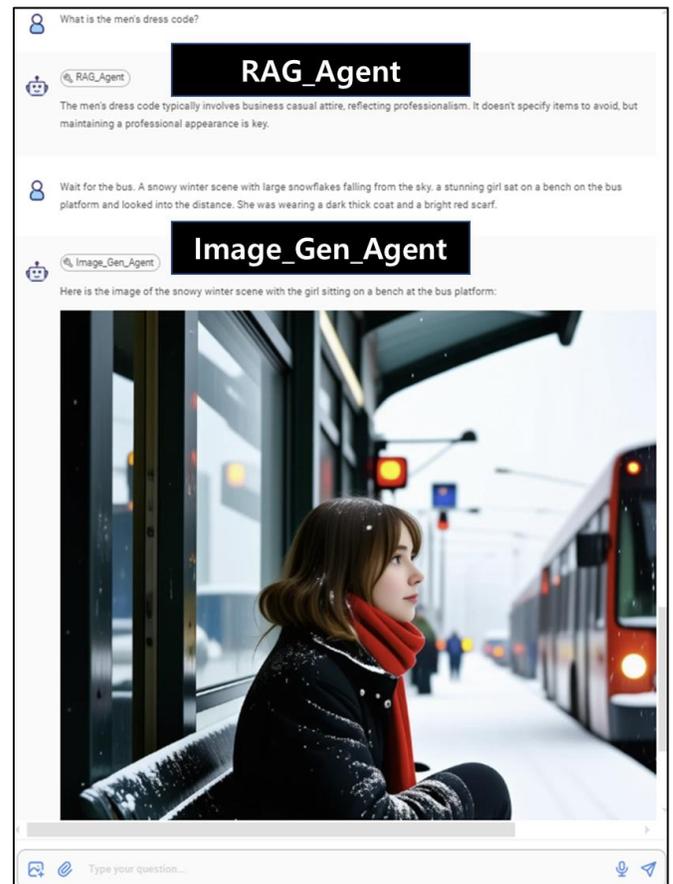

**Fig. (27).** Multi-Agent Integration Implementation Screen

# 6. DISCUSSION AND CONCLUSION

This study proposed a method for designing and implementing a multimodal LLM-based MAS using No-Code tools, demonstrating the feasibility of building AI-driven solutions in enterprises without relying on professional development teams. By leveraging Flowise as a core No-Code platform, the study successfully built a system that handles multimodal data, enhancing the practical applicability of AI. The experimental results validated the system's effectiveness in tasks such as image analysis and code generation, AI-driven image generation, video generation, and RAG-based business query responses.

From a practical perspective, this research holds significance for automating and improving business processes through AI technology. The integration of AI with No-Code platforms significantly lowers the barriers to AI adoption, enabling non-developers to easily incorporate AI capabilities into their workflows. For example, the image analysis and automatic code generation case demonstrated how incomplete code sketches can be transformed into fully functional code, while the image and video generation agent showcased its ability to quickly create creative content, contributing to marketing and branding efforts.

From an academic perspective, this study provides a novel methodology for integrating multimodal LLMs with No-Code-based MAS, offering valuable insights for future research. The framework developed for processing multimodal data and enabling interaction among multiple agents can serve as foundational material for next-generation AI system research. Additionally, the combination of RAG systems with multimodal processing modules introduced a methodology to improve the accuracy and efficiency of document retrieval and response generation.

However, the study has several limitations. The constraints of the No-Code platform limited complex customizations, making it difficult to fully meet domain-specific requirements. The need for more advanced preprocessing and post processing techniques to improve the accuracy of multimodal data handling was also identified. Furthermore, enhancing communication performance and reliability between agents requires further investigation.

## FUTURE RESEARCH DIRECTIONS

Future research should focus on optimizing the performance of agents that handle multimodal data. Enhancements in preprocessing and postprocessing, particularly in OCR and image processing, are necessary to improve text recognition accuracy. Additionally, optimizing communication protocols for seamless data exchange between agents will be essential.

The MAS implemented in this study can be expanded functionally through the integration of additional APIs and modular extensions. Future research could explore incorporating additional agents and new multimodal processing modules to broaden the system's application scope and adapt it to a wider range of business processes. For instance, the integration of complex modules, such as voice recognition, could be considered.

While this study explored the potential of a No-Code-based MAS, further research tailored to the unique needs of various industries is necessary. Developing and evaluating customized agents, data processing techniques, and user interfaces for each industry would be ideal. In particular, research on interactive interfaces aimed at improving user experience will be critical for future advancements.

## AUTHORS' CONTRIBUTION

It is hereby acknowledged that all authors have accepted responsibility for the manuscript's content and consented to its submission. They have meticulously reviewed all results and unanimously approved the final version of the manuscript.

## CONSENT FOR PUBLICATION

Not applicable.

## AVAILABILITY OF DATA AND MATERIALS

The data and supportive information are available within the article.

## FUNDING

None.

## CONFLICT OF INTEREST

The authors declare no conflict of interest, financial or otherwise.

## ACKNOWLEDGEMENTS

Declared none.